\documentclass[sigconf]{acmart}

\usepackage{bm}
\usepackage{balance}
\usepackage{multirow}
\usepackage[table]{xcolor}
\usepackage{makecell} 
\usepackage{xcolor}
\usepackage{float}
\AtBeginDocument{%
  }

\copyrightyear{2026}
\acmYear{2026}
\setcopyright{cc}
\setcctype{by}
\acmConference[WWW '26]{Proceedings of the ACM Web Conference 2026}{April 13--17, 2026}{Dubai, United Arab Emirates}
\acmBooktitle{Proceedings of the ACM Web Conference 2026 (WWW '26), April 13--17, 2026, Dubai, United Arab Emirates}
\acmPrice{}
\acmDOI{10.1145/3774904.3792453}
\acmISBN{979-8-4007-2307-0/2026/04}

\begin{document}


\title{VecFormer: Towards Efficient and Generalizable Graph \\Transformer with Graph Token Attention}





\author{Jingbo Zhou\textsuperscript{$\dagger$}}
\affiliation{%
  \institution{Zhejiang University}
  \city{Hangzhou}
  \state{Zhejiang}
  \country{China}}
\affiliation{%
  \institution{Westlake University}
  \city{Hangzhou}
  \state{Zhejiang}
  \country{China}}
\email{zhoujingbo@westlake.edu.cn}

\author{Jun Xia\textsuperscript{$\dagger$}}
\affiliation{%
  \institution{Hong Kong University of Science and Technology (Guangzhou)}
  \city{Guangzhou}
  \state{Guangdong}
  \country{China}}
\email{junxia@hkust-gz.edu.cn}

\author{Siyuan Li\textsuperscript{$\dagger$}}
\affiliation{%
  \institution{Westlake University}
  \city{Hangzhou}
  \state{Zhejiang}
  \country{China}}
\email{lisiyuan@westlake.edu.cn}

\author{Yunfan Liu}
\affiliation{%
  \institution{Westlake University}
  \city{Hangzhou}
  \state{Zhejiang}
  \country{China}}
\email{liuyunfan@westlake.edu.cn}

\author{Wenjun Wang}
\affiliation{%
  \institution{Tongji University}
  \city{Shanghai}
  \country{China}}
\email{2251275@tongji.edu.cn}

\author{Yufei Huang}
\affiliation{%
  \institution{Westlake University}
  \city{Hangzhou}
  \state{Zhejiang}
  \country{China}}
\email{huangyufei@westlake.edu.cn}

\author{Changxi Chi}
\affiliation{%
  \institution{Westlake University}
  \city{Hangzhou}
  \state{Zhejiang}
  \country{China}}
\email{chichangxi@westlake.edu.cn}

\author{Mutian Hong}
\affiliation{%
  \institution{ShanghaiTech University}
  \city{Shanghai}
  \country{China}}
\email{hongmt2022@shanghaitech.edu.cn}

\author{Zhuoli Ouyang}
\affiliation{%
  \institution{Southern University of Science and Technology}
  \city{Shenzhen}
  \state{Guangdong}
  \country{China}}
\email{ouyangzl2023@mail.sustech.edu.cn}

\author{Shu Wang}
\affiliation{%
  \institution{Jilin University}
  \city{Changchun}
  \state{Jilin}
  \country{China}}
\email{168778017@qq.com}

\author{Zhongqi Wang}
\affiliation{%
  \institution{Jilin University}
  \city{Changchun}
  \state{Jilin}
  \country{China}}
\email{wangzq2024@mails.jlu.edu.cn}

\author{Xingyu Wu}
\affiliation{%
  \institution{Zhejiang University}
  \city{Hangzhou}
  \state{Zhejiang}
  \country{China}}
\email{wuxingyu@zju.edu.cn}

\author{Chang Yu\textsuperscript{$\dagger$}}
\affiliation{%
  \institution{Westlake University}
  \city{Hangzhou}
  \state{Zhejiang}
  \country{China}}
\email{yuchang@westlake.edu.cn}

\author{Stan Z. Li}
\affiliation{%
  \institution{Westlake University}
  \city{Hangzhou}
  \state{Zhejiang}
  \country{China}}
\email{stan.zq.li@westlake.edu.cn}
\authornote{Stan Z. Li is the corresponding author. \\ \textsuperscript{$\dagger$}These authors contributed equally to this work.}





\renewcommand{\shortauthors}{Jingbo Zhou et al.}

\begin{abstract}
Graph Transformer has demonstrated impressive capabilities in the field of graph representation learning. However, existing approaches face two critical challenges: (1) most models suffer from exponentially increasing computational complexity, making it difficult to scale to large graphs; (2) attention mechanisms based on node-level operations limit the flexibility of the model and result in poor generalization performance in out-of-distribution (OOD) scenarios. 
To address these issues, we propose \textbf{VecFormer} (the \textbf{Vec}tor Quantized Graph Trans\textbf{former}), an efficient and highly generalizable model for node classification, particularly under OOD settings. VecFormer adopts a two-stage training paradigm. In the first stage, two codebooks are used to reconstruct the node features and the graph structure, aiming to learn the rich semantic \texttt{Graph Codes}. In the second stage, attention mechanisms are performed at the \texttt{Graph Token} level based on the transformed cross codebook, reducing computational complexity while enhancing the model's generalization capability. Extensive experiments on datasets of various sizes demonstrate that VecFormer outperforms the existing Graph Transformer in both performance and speed.
\end{abstract}





\maketitle

\section{Introduction}
Graph data is prevalent in various domains \cite{DBLP:journals/iotj/XuLJZ23,DBLP:conf/kdd/YingHCEHL18,Strokach2020FastAF,unknown,chi2025grape,cheng2024genohoptionbridginggenenetwork,chengprescribe,jiang2025information}
such as social networks, bioinformatics, multimodality and transportation systems. Graph Neural Networks (GNNs) \cite{kipf2017semisupervisedclassificationgraphconvolutional,veličković2018graphattentionnetworks} learn node representation from graph structure through the message-passing mechanism, achieving significant success in many graph machine learning tasks. However, GNNs still face limitations in capturing global information and long-range dependencies \cite{dwivedi2023longrangegraphbenchmark,wu2022representinglongrangecontextgraph,zhou2024deep}. To address these challenges, Graph Transformers \cite{ying2021transformersreallyperformbad,zhang2020graphbertattentionneededlearning,xing2024moreoverglobalizingproblemgraph} have emerged, drawing inspiration from the success of Transformer in natural language processing. Graph Transformers can effectively extract global features by leveraging the self-attention mechanism to model complex relationships between nodes.

Graph Transformers have shown potential in handling graph-structured data, but they face significant limitations regarding efficiency and generalization in the node classification task. \textbf{First, they typically rely on the self-attention mechanism, which results in a computational complexity of \( O(N^2) \)}, where \( N \) is the number of nodes in the graph. This presents challenges in terms of memory and computational resources when processing large-scale graphs. To address this computational bottleneck, previous works fall into two categories: (1) employing linear attention to reduce the complexity of computing attention through matrix multiplication \cite{wu2023nodeformerscalablegraphstructure,wu2024sgformersimplifyingempoweringtransformers}, and (2) introducing global or partitioned token nodes, enabling each node to interact with these token nodes to obtain global information \cite{fu2024vcrgraphormerminibatchgraphtransformer,chen2023nagphormertokenizedgraphtransformer,zhu2024anchorgtefficientflexibleattention}. However, each of the above methods has its drawbacks. Linear attention-designed mapping functions often struggle to achieve the expressiveness of the self-attention mechanism in most datasets. Introducing global or local token nodes explicitly retains global information, but it becomes challenging to learn finer-grained relationships between nodes. \textbf{Secondly, current methods overly rely on node-level attention learning, which affects their generalizability in out-of-distribution (OOD) scenarios}. Existing approaches perform attention at the node-level, struggling to capture useful patterns due to distributional differences between the training and test sets in the OOD scenarios, which impacts the model's generalization ability. Hence, a natural question emerges: \textit{\textbf{Can we design an efficient and generalizable Graph Transformer?}}

\begin{figure}[htbp]
    \centering
    \includegraphics[width=0.5\textwidth]{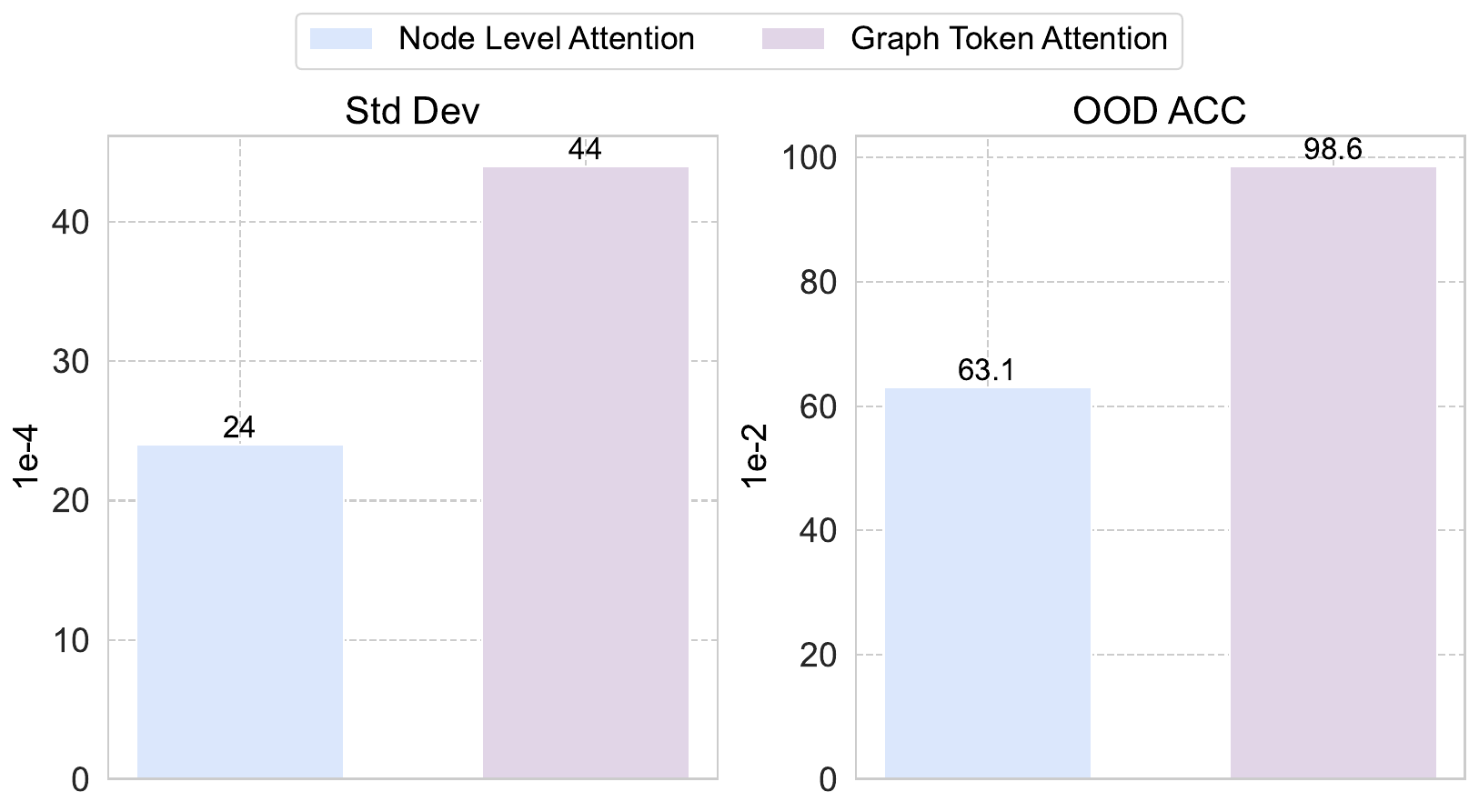} 
    \caption{Standard deviation of attention weights and performance evaluation of Node Level Attention and Graph Token Attention in OOD setting.}
    \label{att_vis} 
\end{figure}

To answer this question, instead of designing a more complex Graph Transformer that combines multiple levels of attention mechanisms, we propose breaking away from the current node-level attention framework and modeling attention at the graph token level, which represents the distribution of graph data.
More specifically, in the VecFormer's first training stage, we introduce a variant of VQVAE\cite{oord2018neuraldiscreterepresentationlearning}, SoftVQ\cite{foldtoken}, to quantize the GNN output to learn the \texttt{Graph Codes} with rich semantic information by reconstructing both the node features and the graph structure. In the second stage, we combine the codes from the transformed structure codebook and feature codebook pairwise to obtain a cross-codebook that constructs \texttt{Graph Tokens}. We then perform cross-attention between the quantized node representations and all \texttt{Graph Tokens} in the cross-codebook. The output is combined with the quantized node representations to form the final output of the model. By introducing soft vector quantization into the Graph Transformer, we effectively compress the information distribution in the graph, reducing the computational complexity. Moreover, we theoretically and experimentally demonstrate the superiority of the graph token attention mechanism, enabling VecFormer to capture beneficial patterns and enhancing the generalization ability in OOD scenarios. We highlight our contributions as follows.

\begin{itemize}
    \item \textit{\underline{Significance}:} We emphasize the importance of attention on general \texttt{Graph Token} that represents the data distribution, to move away from node-level attention, which limits the efficiency and generalization of Graph Transformers.
    \item \textit{\underline{Algorithm}:} We developed a novel algorithm, VecFormer, which uses soft vector quantization to reduce the attention computation complexity on graphs while improving the generalization ability of Graph Transformers in OOD scenarios.
    \item  \textit{\underline{Experiments}:} We experimentally show that the proposed method outperforms other Graph Transformers in terms of node classification task, OOD scenarios, and efficiency.
\end{itemize}

\section{Related Work}

\subsection{Graph Transformer}


Graph Transformers (GTs) have emerged as a powerful approach for graph representation learning, leveraging attention mechanisms to capture complex relationships within graph-structured data. Various methodologies have been proposed to enhance their efficiency and scalability. NodeFormer \cite{wu2023nodeformerscalablegraphstructure} introduces a kernelized Gumbel-Softmax operator to facilitate efficient all-pair message passing with linear complexity, making it suitable for large-scale node classification tasks. Similarly, SGFormer \cite{wu2024sgformersimplifyingempoweringtransformers}  simplifies the architecture by employing a single-layer global attention mechanism, achieving significant inference acceleration without complex preprocessing. GOAT \cite{pmlr-v202-kong23a} enhances the adaptability of node relationships by allowing each node to attend to all others in the graph, effectively addressing homophily and heterophily in node classification. GraphGPS \cite{rampášek2023recipegeneralpowerfulscalable}  presents a modular framework that integrates local message passing with global attention, maintaining linear complexity while achieving state-of-the-art performance across various tasks. Exphormer \cite{shirzad2023exphormersparsetransformersgraphs} leverages expander graphs and sparse attention to enhance scalability, achieving linear time complexity while maintaining competitive accuracy. Polynormer 
\cite{deng2024polynormerpolynomialexpressivegraphtransformer} further advances the field by employing a local-to-global attention mechanism that learns high-degree polynomial functions from node features efficiently. Lastly, NAGphormer \cite{chen2023nagphormertokenizedgraphtransformer} innovatively treats nodes as sequences of tokens through its Hop2Token module, enabling scalable multi-hop neighborhood aggregation.
In contrast to these node-level attention methods, VecFormer applies graph token attention using vector quantization techniques.

\subsection{OOD Generalization on Graph}
 
Out-of-distribution (OOD) generalization in Graph Neural Networks (GNNs) is a critical area of research, as  GNNs often struggle when the test data distribution differs from the training data \cite{yang2025nodereg,yang2025bounded,yangharnessing}. Various approaches have been proposed to tackle this challenge. Invariant Risk Minimization (IRM) \cite{arjovsky2020invariantriskminimization} aims to capture invariant correlations across multiple training distributions, providing a framework for learning robust representations. Deep Coral \cite{sun2016deepcoralcorrelationalignment} extends the original Coral method to learn nonlinear transformations that align layer activation correlations, enhancing model adaptability to distribution shifts. Domain-Adversarial Neural Networks (DANN) \cite{ganin2016domainadversarialtrainingneuralnetworks} introduce domain-adapted representations, bridging the gap between similar yet distinct training and testing distributions. Group Distributionally Robust Optimization (GroupDRO) \cite{sagawa2020distributionallyrobustneuralnetworks} emphasizes regularization for worst-group generalization, achieving improved accuracy by combining group DRO with increased regularization. Mixup \cite{zhang2018mixupempiricalriskminimization} enhances the training dataset by interpolating between observed samples, thereby promoting robustness against distributional discrepancies. The SR-GNN \cite{zhu2021shiftrobustgnnsovercominglimitations} model addresses these discrepancies specifically in graph contexts by adjusting GNN architectures to account for shifts between labeled nodes in the training set and the broader dataset. EERM \cite{wu2024handlingdistributionshiftsgraphs} employs adversarial training based on invariance principles to facilitate environmental exploration. Recent advancements like CaNet \cite{wu2024graphoutofdistributiongeneralizationcausal} focus on preserving causal invariance among nodes to bolster OOD generalization capabilities.
VecFormer narrows the distributional discrepancy between the training and test sets through the SoftVQ technique, thereby enhancing the generalization ability of the attention mechanism. 

\subsection{Vector Quantization in Graph Neural Networks}

Recent advances in vector quantization (VQ) techniques have significantly advanced graph representation learning across biological domains. The foundational work of VQ-GNN\cite{ding2021vqgnnuniversalframeworkscale} introduced a learnable codebook mechanism to map continuous node embeddings into discrete semantic spaces, employing hierarchical quantization and attention-based code assignment to balance structural preservation with computational efficiency in large-scale graph processing. Mole-BERT\cite{xia2023molebert} revolutionized molecular graph pre-training through a type-constrained VQ-VAE architecture, where specialized codebook partitions for different atom types (e.g., dedicated ranges for carbon/nitrogen/oxygen) prevent cross-type interference while enabling context-aware discretization of atomic environments. This framework synergized masked atom modeling with triplet contrastive learning to capture both local reconstructive and global relational patterns.
The Prompt-DDG\cite{wu2024learningpredictmutationeffects} established a hierarchical prompt codebook system that decomposes 3D microenvironments into residue-type, angular statistics, and local conformation subspaces, enabling efficient mutation effect prediction through masked microenvironment modeling. Concurrently, the MAPE-PPI\cite{wu2024mapeppieffectiveefficientproteinprotein} framework achieved state-of-the-art performance by integrating VQ-derived discrete prompts that encode microenvironmental variations with minimal parameter overhead ($<$5\%). VQGraph\cite{yang2024vqgraphrethinkinggraphrepresentation} propose a dual-channel codebook architecture to bridge GNNs and MLPs through discrete-continuous hybrid representations. This design enables hard index selection from one codebook channel while performing soft embedding retrieval from another, effectively aligning GNN's structural awareness with MLP's inference speed. In this work, we focus on enhancing the  GT's computational efficiency and the generalization ability in OOD scenarios through vector quantization techniques.

\section{Preliminary}
A connected undirected graph, denoted by $\mathcal{G}=(\mathcal{V}, \mathcal{E})$, consists of the set of $N$ nodes $\mathcal{V}=\left\{v_1, v_2, \ldots, v_N\right\}$ and the set of edges $\mathcal{E} \subseteq \mathcal{V} \times \mathcal{V}$. The node features are represented in the matrix $\boldsymbol{X} \in \mathbb{R}^{N \times d}$, where $\boldsymbol{x}_i$ denotes the feature of node $v_i$ and $d$ represents the dimension of the feature. $\boldsymbol{A} \in\{0,1\}^{N \times N}$ denotes the adjacency matrix, where $\boldsymbol{A}_{i j}=1$ signifies the presence of an edge between nodes $v_i$ and $v_j$ in the graph. The adjacency matrix with self-loop is represented as $\hat{\boldsymbol{A}}=\boldsymbol{A}+\boldsymbol{I}$. Additionally, we use $\boldsymbol{H}$ to denote the output of the GNN layer, representing the new representation matrix of node features.

\subsection{Graph Transformer}

Graph Transformers leverage a global attention mechanism to enable each node to interact with every other node in the graph. The node features \( \bm{H} \) are updated as follows:
\begin{equation}
\bm{Q} = \bm{H} \bm{W}_Q, \quad \bm{K} = \bm{H} \bm{W}_K, \quad \bm{V} = \bm{H} \bm{W}_V,
\end{equation}
\begin{equation}
\text{Attention}(\bm{H}) = \text{Softmax}\left(\frac{\bm{Q} \bm{K}^\top}{\sqrt{d}}\right) \bm{V},
\end{equation}
where \( N \) is the number of nodes, \( \bm{W}_Q, \bm{W}_K, \bm{W}_V \in \mathbb{R}^{d \times d} \) are learnable projection matrices. The attention scores are calculated using the scaled dot-product of \( \bm{Q} \) and \( \bm{K} \), normalized with the softmax function. These scores are then used to aggregate information from all nodes into \( \bm{V} \), enabling effective global information propagation.

\subsection{Vector Quantization}

\textbf{Vanilla VQ}  
The VQ-VAE model, introduced by \citet{oord2018neuraldiscreterepresentationlearning}, is designed to model continuous data distributions, such as images, audio, and video, by encoding them into discrete latent variables. The encoder maps the input \(x\) to a continuous latent vector \(\mathbf{E}(x)\), which is then quantized to the nearest codebook vector \(e_k\) based on a minimum distance criterion:
\begin{equation}
\text{Quantize}(\mathbf{E}(x)) = e_k, \quad k = \arg \min_j \|\mathbf{E}(x) - e_j\|.
\end{equation}
The model is trained using the reconstruction loss, which measures the difference between the input \(x\) and the decoder output \(\mathbf{D}(e)\), alongside two additional terms: the codebook loss, which ensures that the selected codebook vector is close to the encoder output, and the commitment loss, which stabilizes the encoder by encouraging \(\mathbf{E}(x)\) to stay close to the chosen codebook vector. The total loss function is expressed as:
\begin{small}
\begin{equation}
\mathcal{L}(x, \mathbf{D}(e)) = \|x - \mathbf{D}(e)\|_2^2 + \|\text{sg}[\mathbf{E}(x)] - e\|_2^2 + \eta \|\text{sg}[e] - \mathbf{E}(x)\|_2^2,
\end{equation}
\end{small}

where \(\text{sg}\) denotes the stop-gradient operation, and \(\eta\) is a hyperparameter that controls the relative learning speed.

\begin{figure*}[htbp]
    \centering
    \includegraphics[width=1.02\textwidth]{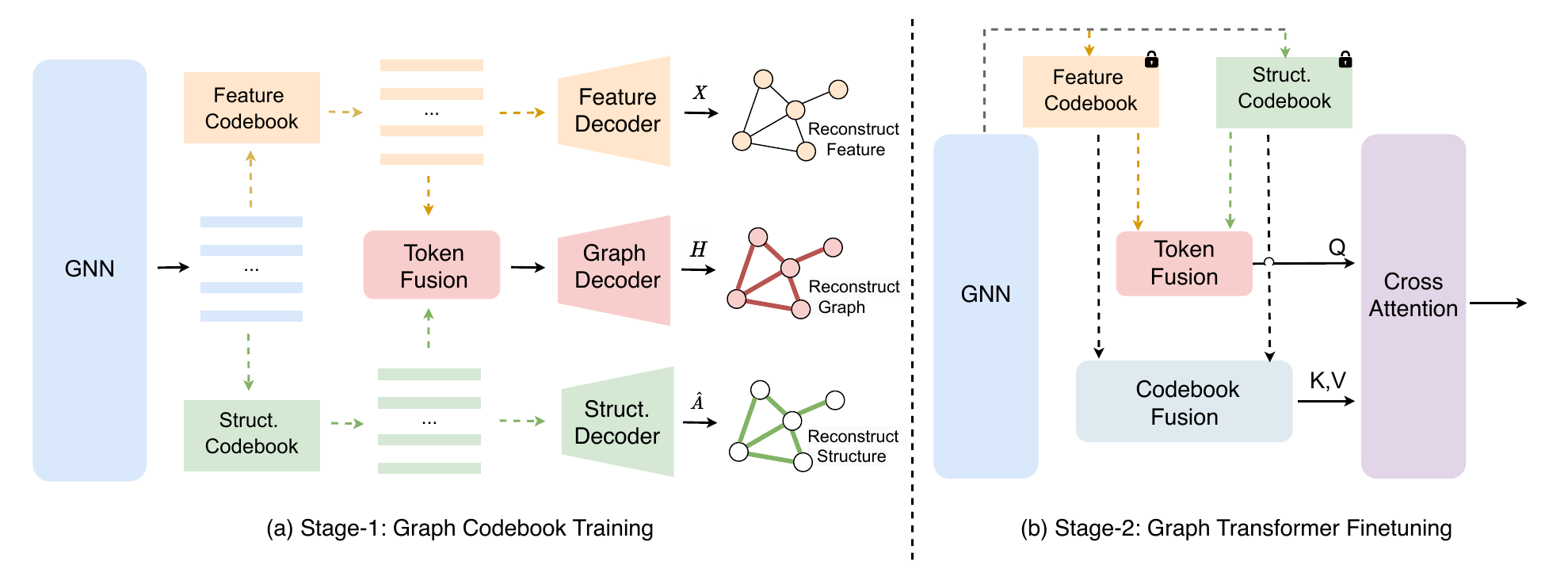} 
        \caption{Illustration of the VecFormer's architecture and the training process.}
    \label{vecformer} 
\end{figure*}

\section{VecFormer}

In this section, we provide a detailed introduction to VecFormer. The core idea of VecFormer is to discretely compress the representational space of feature and structural distributions on graph, learn semantically rich \texttt{Graph Codes}, and apply them in a Transformer with a graph token attention mechanism. This approach enhances the model's generalization ability while reducing computational efficiency. As shown in Figure \ref{vecformer}, the training of this model consists of two stages: \textbf{Graph Codebook Training} and \textbf{Graph Transformer Finetuning}.
The basic notations related to VecFormer and their meanings are listed in Table \ref{notation_1}.

\begin{table}[h]
\centering
\caption{Basic notation and their meanings.}
\begin{tabular}{cc}
\toprule
\textbf{Notation} & \textbf{Meaning} \\ \midrule
$\mathcal{G}$ & Attribute Graph \\ 
$N$ & Number of nodes \\ 
$A$ & Adjacency matrix \\ 
$T$ & Temperature parameter \\ 
${X} \in \mathbb{R}^{n \times d}$ & Feature matrix \\ 
$H \in \mathbb{R}^{n \times d'}$ & GNN's output matrix \\ 
\( F_C =   \{fc_z\}_{z=1}^{m} \) & Feature Codebook \\ 
\( S_C = \{sc_z\}_{z=1}^{n} \) &
Structure Codebook \\
$G$ & The nodes' corresponding Graph Token\\
$F_T, S_T,G_T$ & Feature, Structure and Graph Token List \\ 
\bottomrule
\end{tabular}
\label{notation_1}
\end{table}

\vspace{-1em}
\subsection{Graph Codebook Training}

In the Graph Codebook Training stage, we employ two codebooks: a Feature Codebook and a Structure Codebook. Using the SoftVQ approach, we define self-supervised tasks including feature reconstruction, structure reconstruction, and joint reconstruction of graph features. These tasks aim to learn semantically rich \texttt{Graph Codes}, consisting of \texttt{Feature Codes} and \texttt{Structure Codes}. For each node, the corresponding \texttt{Feature Token} and \texttt{Structure Token} are obtained through a linear combination of codes from the Feature Codebook and Structure Codebook using the SoftVQ method. The \texttt{Graph Token} for the node is generated by fusing the \texttt{Feature Token} and \texttt{Structure Token} through a Token Fusion module. This process effectively quantizes the node and derives its corresponding \texttt{Graph Token}, providing the foundation for the Graph Token-level attention mechanism in the Graph Transformer Finetuning stage.

\textbf{Soft VQ} Vanilla VQ is prone to codebook collapse during training, where the model quantizes all samples to a few codebook vectors, leading to training failure. To address this, we introduce the SoftVQ to improve vector quantization by introducing a differentiable attention mechanism. Instead of selecting a single closest codebook vector, SoftVQ computes a set of attention weights \(a_{ij}\) that quantify the contribution of each codebook vector \({e}_j\) to the input feature \({h}_i\).  The final output \(\hat{{h}}_i\) is computed as a weighted sum of the codebook vectors:
\begin{equation}
\small a_{ij} = \frac{\exp\left({{h}_i^T {e}_j}{/T}\right)}{\sum_{j=1}^{m} \exp\left({{h}_i^T {e}_j}{/T}\right)}
, \quad \hat{{h}}_i = \sum_{j=1}^{m} a_{ij}{e}_j,
\end{equation}
where \(T\) is a temperature parameter that controls the sharpness of the attention distribution, with higher values leading to more even contributions from the codebook vectors. This approach allows for a smoother, more flexible representation compared to hard quantization.

\textbf{Using Feature and Structure Codebooks to Learn Graph Data Distribution} VQ-VAE has been widely applied in computer vision (CV). Unlike images, which are Euclidean data, graphs are non-Euclidean data with two dimensions: features and structure. In VecFormer, we employ a Feature Codebook and a Structure Codebook to learn the feature and structural distributions of graph data.
Given a graph $\mathcal{G}$ with nodes \( \mathcal{V} = \{v_1, v_2, \dots, v_N\} \), each node has corresponding features \( \{x_1, x_2, \dots, x_N\} \). Let the \textbf{Feature Codebook} be \( \{fc_z\}_{z=1}^{m} \) and the \textbf{Structure Codebook} be \( \{sc_z\}_{z=1}^{n} \).  For each node, the corresponding \texttt{Feature Token} and \texttt{Structure Token} are obtained through an attention-weighted linear combination of codes from the Feature Codebook and Structure Codebook. The quantization process can be formally expressed as follows:
\begin{equation}
\small f_i = \text{SoftVQ}(h_i, \{fc_z\}_{z=1}^{m}),  \quad s_i = \text{SoftVQ}(h_i, \{sc_z\}_{z=1}^{n}),
\end{equation}
where \( h_i \) is the node representation output by the GNN, and \( f_i \) and \( s_i \) are the \texttt{Feature Token} and \texttt{Structure Token} obtained through quantization. In VecFormer, the GNN encoder can be any Message Passing Neural Network (MPNN) framework. \textbf{It is important to note that graph features and structures are not entirely orthogonal.} For instance, nodes of different class often have distinct neighborhood structures, necessitating the joint modeling of feature and structure distributions. Furthermore, to obtain an overall quantized representation of a node, rather than independent \texttt{Feature Token} and \texttt{Structure Token}, we employ a \textbf{Token Fusion} module to combine the two tokens into a \texttt{Graph Token}. The fusion process is as follows:
\begin{equation}
\small g_i = \alpha_1 \cdot f_i + \alpha_2 \cdot s_i, \quad  (\alpha_1, \alpha_2) = \varphi_w(f_i, s_i) ,
\end{equation}
where \( \varphi_w \) is a linear transformation, \( g_i \) is the \texttt{Graph Token} obtained by fusing the \texttt{Feature Token} \( f_i \) and the \texttt{Structure Token} \( s_i \). We optimize the codebooks using these three types of tokens.

\textbf{Graph Codebook Optimization} To obtain semantically rich \texttt{Graph Codes}, we optimize the Feature Codebook and Structure Codebook through self-supervised tasks involving the reconstruction of node features, graph structure, and graph features. The loss for optimizing the \texttt{Graph Codes} in codebooks is as follows:
\begin{equation}
\small
   \begin{aligned}
    \mathcal{L}_{rec} &= \underbrace{\frac{1}{n}\sum_{i=1}^{n}\left(1-\frac{{x}_{i}{}^{T}\hat{{x}}_{i}}{\|{x}_{i}\|\cdot\|{\hat{x}}_{i}\|}\right)^{\gamma_f}}_{\text{Feature Reconstruction}}+\underbrace{\left\|\boldsymbol{A}-\sigma(\hat{\boldsymbol{Y}}\cdot\hat{\boldsymbol{Y}}^{T})\right\|_{2}^{2}}_{\text{Structure Reconstruction}} \\
    &+ \underbrace{\frac{1}{n}\sum_{i=1}^{n}\left(1-\frac{{h}_{i}{}^{T}\hat{{h}}_{i}}{\|{h}_{i}\|\cdot\|{\hat{h}}_{i}\|}\right)^{\gamma_g}}_{\text{Graph Reconstruction} },
\end{aligned} 
\end{equation}
where \(\hat{x_i}\) represents the node features reconstructed from the \texttt{Feature Token} through the \textbf{Feature Decoder}, while $\boldsymbol{\hat{Y}}$ is the matrix obtained from the \texttt{Structure Token} through the \textbf{Structure Decoder}, which is used to reconstruct the adjacency matrix \(A\). \(\hat{h_i}\) denotes the GNN output reconstructed from the \texttt{Graph Token} via the \textbf{Graph Decoder}, and \(\sigma\) refers to the Sigmoid function. All of the above decoders are implemented using the MLP. Since MSE is sensitive to vector norms and dimensions, we use the scaled cosine error adopted by GraphMAE \cite{hou2022graphmaeselfsupervisedmaskedgraph} for feature and graph reconstruction. \(\gamma_f\) and \(\gamma_g\) are scaling factors. Through the graph feature reconstruction task, we acquire semantically rich feature and structure codebooks that enhance Graph Transformer Finetuning.

\subsection{Graph Transformer Finetuning}
In the Graph Transformer Finetuning phase, we utilize the pretrained codebooks to construct a context-aware \texttt{Graph Token List}, which functions as both the key and value for the cross-attention mechanism. This enables the nodes' graph tokens to capture global graph information during attention interactions with the \texttt{Graph Token List}.

\textbf{Constructing Context-aware Graph Token List} Given the Feature Codebook \( \boldsymbol{F_C} =   \{fc_z\}_{z=1}^{m}  \) and the Structure Codebook \( \boldsymbol{S_C} = \{sc_z\}_{z=1}^{n} \) obtained in the Graph Codebook Training phase, the process for constructing the \texttt{Graph Token List} via \textbf{Codebook Fusion} module is as follows:
\begin{equation}
\boldsymbol{F_T} = (\boldsymbol{F_C}^T \boldsymbol{W_F})^T, \quad
\boldsymbol{S_T} = (\boldsymbol{S_C}^T \boldsymbol{W_S})^T, 
\end{equation}
\begin{equation}
\boldsymbol{G_T} = \text{TokenFusion}(\boldsymbol{F_T}, \boldsymbol{S_T}),
\end{equation}
where \( \boldsymbol{F_T} \) and \( \boldsymbol{S_T} \) are the \texttt{Feature} and \texttt{Structure Token List}, \( \boldsymbol{W_F} \in \mathbb{R}^{m \times N_f} \) and \( \boldsymbol{W_S} \in \mathbb{R}^{n \times N_s} \) are linear projections. Since the \texttt{Feature} and \texttt{Structure Token} are linear combinations of the codes from codebooks that compress the graph distribution,  we set learnable parameters $\boldsymbol{W_F}$ and $\boldsymbol{W_S}$ to transform the codebooks into a context-aware Token List. \( N_f \) and \( N_s \) denote the number of \texttt{Feature Tokens} and \texttt{Structure Tokens}. Then, we fuse the tokens pairwise from the \texttt{Feature Token} and \texttt{Structure Token List} using the \textbf{Token Fusion} module, resulting in the context-aware \texttt{Graph Token List} with the size of \( N_f  N_s \).

\textbf{Cross-Attention}
The nodes' corresponding \texttt{Graph Token} matrix \( \bm{G} \) serves as the query for the attention mechanism. The \texttt{Graph Token List} \( \bm{G_T} \), constructed through the codebooks, functions as both the key and the value. The final output of the model is computed as follows:
\begin{equation}
\bm{Q} = \bm{G} \bm{W}_Q, \quad \bm{K} = \bm{G_T} \bm{W}_K, \quad \bm{V} = \bm{G_T} \bm{W}_V,
\end{equation}
\begin{equation}
\bm{Z} = \text{Softmax}\left( \frac{\bm{Q} \bm{K}^\top}{\sqrt{d}} \right) \bm{V} + \bm{G},
\end{equation}
where \( \bm{Z} \) represents the model's output, \( \bm{W}_Q, \bm{W}_K, \bm{W}_V \in \mathbb{R}^{d \times d} \) are the learnable weight matrices. By applying attention between the \texttt{Graph Token} for node and the context-aware \texttt{Graph Token List} for the graph, the node's representation incorporates global information. The residual connection ensures that the output retains local information. The model is trained using a classification loss.

\subsection{Model Analysis}
\label{model_analysis}
In this section, we introduce how VecFormer reduces computational complexity and enhances the generalization ability of the attention mechanism by employing Graph Token Attention, as opposed to the node-level attention mechanism.

\textbf{Model Computational Efficiency} The computational complexity of node-level attention is \( O(N^2) \). In VecFormer, we reduce the computational complexity to \( O(NN_fN_s) \) by constructing a context-aware \texttt{Graph Token List}, which decreases the token count to \( N_f N_s \), where \( N_f \) and \( N_s \) are typically small. This \( N_f  N_s \) is significantly smaller than \( N \), which can reach millions of nodes, and remains unchanged as the graph size increases, achieving a linear computational complexity of \( O(N) \). It is worth noting that both VecFormer training stages have linear complexity, and since the model converges quickly in the first stage, the overall training time is approximately equal to the training time of the second stage. In the meantime, unlike other linear complexity GTs, VecFormer reduces computational complexity by minimizing the number of tokens requiring attention, which is orthogonal to methods that modify the attention mechanism. Furthermore, VecFormer's attention architecture is simple, does not rely on complex kernel functions, is compatible with widely used attention mechanisms, and is more easily accelerated by hardware.

\begin{table*}[htbp]
  \caption{Node classification performance (\%) of VecFormer and three types of GNNs.  The best results are
in bold, and the second-best results are underlined.}
  \label{homo_dataset}
  \centering
  \resizebox{0.95\linewidth}{!}{
    \begin{tabular}{cccccccc}
      \toprule
      \text{Category} & \text{Model} & \text{Pubmed} & \text{CoraFull} & \text{Computer} & \text{Photo} & \text{CS} & \text{Physics} \\
      \midrule
       & \text{GCN}           & 87.36  $\pm$  0.35  & 67.44  $\pm$  0.64    & 91.08  $\pm$  0.62    & 94.44  $\pm$  0.57  & 93.40  $\pm$  0.24 & 96.53  $\pm$  0.14 \\
      GNN & \text{GAT}           & 86.12 $\pm$ 0.44  & 66.74 $\pm$ 0.75    & 91.31 $\pm$ 0.45    & 94.46 $\pm$ 0.69  & 93.70 $\pm$ 0.29 & 96.47 $\pm$ 0.19 \\
      & \text{SAGE}          & 87.83 $\pm$ 0.34  & 68.59 $\pm$ 0.80    & 90.29 $\pm$ 0.68    & 94.03 $\pm$ 0.69  & 93.04 $\pm$ 0.26 & 96.29 $\pm$ 0.17 \\
    \midrule
      & \text{GOAT}          & 89.15 $\pm$ 0.39  & 53.30 $\pm$ 3.89    & 88.95 $\pm$ 1.00    & 92.11 $\pm$ 0.95  & 89.96 $\pm$ 0.63 & 96.27 $\pm$ 0.30 \\
      & \text{GraphGPS}           & 87.72 $\pm$ 0.30  & 59.11 $\pm$ 2.22    & 87.62 $\pm$ 0.68    & 94.18 $\pm$ 0.64  & \underline{95.42 $\pm$ 0.28} & \underline{97.05 $\pm$ 0.18} \\
      & \text{NodeFormer}    & 81.87 $\pm$ 2.24  & 60.14 $\pm$ 2.97    & 85.69 $\pm$ 1.25    & 93.82 $\pm$ 0.97  & 95.07 $\pm$ 0.39 & 96.80 $\pm$ 0.63 \\
 GT & \text{SGFormer}      & 90.37 $\pm$ 0.20  & \underline{70.11 $\pm$ 0.28}    & 90.75 $\pm$ 0.34    & 94.81 $\pm$ 0.43  & 95.16 $\pm$ 0.28 &96.63 $\pm$ 0.14 \\
       & \text{Exphormer}     & 89.06 $\pm$ 0.24  & 63.70 $\pm$ 0.71    & 88.85 $\pm$ 0.26    & 94.10 $\pm$ 0.31  & 95.03 $\pm$ 0.47 & 96.93 $\pm$ 0.07 \\
        & \text{Polynormer}    & \underline{90.44 $\pm$ 0.21}  & 66.54 $\pm$ 0.89    & \underline{91.92 $\pm$ 0.25}  & \underline{95.44 $\pm$ 0.24} & 95.38  $\pm$  0.32 &96.89  $\pm$  0.06 \\
    &   \text{NAGphormer}    & 88.65 $\pm$ 0.42  & 65.51 $\pm$ 0.64    & 89.71 $\pm$ 0.60    & 95.03 $\pm$ 0.77  & 95.23 $\pm$ 0.24 & 96.86 $\pm$ 0.21 \\
\midrule
VQ GNN & \text{VQGraph}     &     89.05 $\pm$ 0.32 & 63.63 $\pm$ 0.84 & 85.00 $\pm$ 1.04 & 86.56 $\pm$ 2.39 & 92.36 $\pm$ 5.87 & 96.46 $\pm$ 0.20
\\
\midrule
\rowcolor[gray]{0.9}
 VQ GT & \text{VecFormer}  & \textbf{90.47 $\pm$ 0.28}& \textbf{72.14 $\pm$ 0.67}& \textbf{92.51 $\pm$ 0.54}& \textbf{95.84 $\pm$ 0.41}& \textbf{95.54 $\pm$ 0.34}& \textbf{97.17 $\pm$ 0.24}\\
       \bottomrule
    \end{tabular}
  }
\end{table*}

\textbf{Model Generalization Ability} In the Graph OOD scenarios, when performing self-attention mechanisms, \( q \) and \( k \) often come from distributions with significant differences. Due to differences in node distributions in the Graph OOD scenario and the fact that the OOD set is not included in the training set, resulting in the low similarity between \(q \) and \(k\), the value of \( q \cdot k \) is typically close to or equal to 0. When the lower bound of \( q \cdot k \) is 0, the attention coefficients are computed as follows:
\begin{equation}
    \text{Softmax}(qK) \sim U,
\end{equation}
where \(U\) is uniform distribution. The importance of \(V\) remains constant across all values of $Q$, making it difficult for the attention mechanism to learn meaningful patterns. In VecFormer, since the \texttt{Graph Token} is a linear combination of the pretrained \texttt{Graph Codes}, we can express \(q\) and \(k\) as follows:
\begin{equation}
q = \sum_{i=1}^{m+n} \alpha_i c_i , \quad k = \sum_{i=1}^{m+n} \beta_i c_i,
\end{equation}
where \(\alpha\), \(\beta\) denote the weight coefficients, and  $c$ represents the Feature and Structure Code. When the code is orthogonal to any other code,   \(q \cdot k\)  can be computed as
\begin{equation}
    q \cdot k \geq \sum_{i=1}^{m+n} \alpha_i \beta_i ||c_i||^2 > 0 .
\end{equation}
The lower bound of \( q \cdot k \) is no longer zero. VecFormer is capable of learning diverse attention patterns by adjusting \( \alpha \) and \( \beta \), thereby enhancing the model's generalization ability.
To further validate our point, we compare the standard deviation of learned attention weights and the performance of VecFormer and the vanilla  Transformer on OOD datasets.
As shown in Figure \ref{att_vis}, in the OOD scenario, VecFormer exhibits a higher standard deviation in attention weights, suggesting more diverse attention patterns.  Furthermore, VecFormer consistently achieves superior performance compared to the vanilla Transformer on OOD test sets.  These results align well with our underlying explanation of the method's effectiveness, providing empirical evidence that supports the validity of our design rationale.


\section{Experiments}

In this section, we comprehensively evaluate our proposed VecFormer by addressing the following research questions \textbf{(RQs)}:

\begin{itemize}
    \item \textbf{RQ1: Superiority.} Does VecFormer outperform other Graph Transformers on node classification tasks?
    \item \textbf{RQ2: Generalization.} Does our proposed method achieve greater improvements in graph out-of-distribution (OOD) scenarios?
    \item \textbf{RQ3: Scalability.} How does VecFormer perform in large graphs in terms of effectiveness and efficiency?
    \item \textbf{RQ4: Effectiveness.} Is the application of vector quantization techniques effective in Graph Transformers?
\end{itemize}

\begin{table*}[htbp]
  \caption{Results of classification accuracy (\%) on OOD datasets. The best results are in bold, and the second-best results are underlined. $\Delta_{OODM}$ and $\Delta_{GT}$ represent the performance improvement of VecFormer compared to the best-performing OOD method and Graph Transformer on each dataset, respectively. }
  \label{ood_result}
  \centering
  \resizebox{0.95\linewidth}{!}{
    \begin{tabular}{ccc|cc|cccc}
\toprule
\multirow{2}{*}{\text{Model}} & \multicolumn{2}{c}{\text{Cora}} & \multicolumn{2}{c}{\text{Citeseer}} &  \multicolumn{4}{c}{\text{Twitch}}\\  
& \text{OOD} & \text{ID} & \text{OOD} & \text{ID}  & \text{OOD1}& \text{OOD2}& \text{OOD3}&\text{ID}  \\
\midrule
ERM & 91.10 $\pm$ 2.26 & 95.57 $\pm$ 0.40 & 82.60 $\pm$ 0.51 & 89.02 $\pm$ 0.32  & 65.67 $\pm$ 0.02 & 52.00 $\pm$ 0.10 & 61.85 $\pm$ 0.05 &75.75 $\pm$ 0.15 \\
IRM & 91.63 $\pm$ 1.27 & 95.72 $\pm$ 0.31 & 82.73 $\pm$ 0.37 & 89.11 $\pm$ 0.36  & 67.27 $\pm$ 0.19 & 52.85 $\pm$ 0.15 & 62.40 $\pm$ 0.24 &75.30 $\pm$ 0.09 \\
Coral & 91.82 $\pm$ 1.30 & 95.74 $\pm$ 0.39 & 82.44 $\pm$ 0.58 & 89.05 $\pm$ 0.37  & 67.12 $\pm$ 0.03 & 52.61 $\pm$ 0.01 & 63.41 $\pm$ 0.01 &75.20 $\pm$ 0.01 \\
DANN & 92.40 $\pm$ 2.05 & 95.66 $\pm$ 0.28 & 82.49 $\pm$ 0.67 & 89.02 $\pm$ 0.31  & 66.59 $\pm$ 0.38 & 52.88 $\pm$ 0.12 & 62.47 $\pm$ 0.32 &75.82 $\pm$ 0.27 \\
GroupDRO & 90.54 $\pm$ 0.94 & 95.38 $\pm$ 0.23 & 82.64 $\pm$ 0.61 & 89.13 $\pm$ 0.27  & 67.41 $\pm$ 0.04 & 52.99 $\pm$ 0.08 & 62.29 $\pm$ 0.03 &75.74 $\pm$ 0.02 \\
Mixup & 92.94 $\pm$ 1.21 & 94.66 $\pm$ 0.10 & 82.77 $\pm$ 0.30 & 89.05 $\pm$ 0.05  & 65.58 $\pm$ 0.13 & 52.04 $\pm$ 0.04 & 61.75 $\pm$ 0.13 &75.72 $\pm$ 0.07 \\
SRGNN & 91.77 $\pm$ 2.43 & 95.36 $\pm$ 0.24 & 82.72 $\pm$ 0.35 & 89.10 $\pm$ 0.15  & 66.17 $\pm$ 0.03 & 52.84 $\pm$ 0.04 & 62.07 $\pm$ 0.04 &75.45 $\pm$ 0.03 \\
EERM & 91.80 $\pm$ 0.73 & 91.37 $\pm$ 0.30 & 74.07 $\pm$ 0.75 & 83.53 $\pm$ 0.56  & 66.80 $\pm$ 0.46 & 52.39 $\pm$ 0.20 & 62.07 $\pm$ 0.68 &75.19 $\pm$ 0.50 \\
CaNet & \underline{97.30 $\pm$ 0.25} & 95.94 $\pm$ 0.29 & \underline{95.33 $\pm$ 0.33} & 89.57 $\pm$ 0.65  & \underline{68.08 $\pm$ 0.19} & 53.49 $\pm$ 0.14 & \underline{63.76 $\pm$ 0.17} & \underline{76.14 $\pm$ 0.07} \\
\midrule
 NodeFormer& 66.81  $\pm$  1.80& 88.48  $\pm$  3.27& 60.00  $\pm$  1.09& 81.15  $\pm$  0.20& 61.75  $\pm$  0.45& 50.68  $\pm$  0.06& 56.99  $\pm$  0.10&57.80  $\pm$  0.13\\
 SGFormer& 90.18  $\pm$  0.95& 96.29  $\pm$  0.15 & 78.26  $\pm$  0.69& \underline{ 93.95  $\pm$  0.10}& 65.11  $\pm$  0.13& 53.63  $\pm$  0.11& 62.20  $\pm$  0.06&67.90  $\pm$  0.12\\
 Exphormer& 88.76  $\pm$  0.88& \underline{96.35  $\pm$  0.19}& 80.54  $\pm$  1.25& 92.14  $\pm$  0.10& 66.22  $\pm$  0.43& 53.69  $\pm$  0.15& 63.26  $\pm$  0.30&70.54  $\pm$  0.44\\
 Polynormer& 87.00  $\pm$  1.32& 94.96  $\pm$  0.31& 73.95  $\pm$  1.27& 91.95  $\pm$  0.23& 66.07  $\pm$  0.13& \underline{53.90  $\pm$  0.16}& 62.93  $\pm$  0.07&74.48  $\pm$  0.30\\
 NAGphormer& 84.51  $\pm$  2.29& 84.35  $\pm$  2.35& 65.91  $\pm$  5.21& 84.86 ± 1.20& 62.85  $\pm$  2.88& 52.10  $\pm$  1.41& 59.01  $\pm$  3.37&71.42  $\pm$  1.61\\
\rowcolor[gray]{0.9}
  VecFormer& \textbf{98.85 $\pm$ 0.08}& \textbf{97.91 $\pm$ 0.14} & \textbf{97.45 $\pm$ 0.04}& \textbf{94.94 $\pm$ 0.20} & \textbf{68.14 $\pm$ 0.14}& \textbf{54.08 $\pm$ 0.16}& \textbf{64.27 $\pm$ 0.18}& \textbf{76.51$\pm$ 0.09}\\
 \midrule
$\Delta_{OODM}$ & +1.55\%   & +1.97\%  &  +2.12\%   & +5.37\%   & +0.06\%   & +0.59\%   & +0.51\%   & +0.37\%   \\
$\Delta_{GT}$ & +8.67\%   & +1.56\%   &  +16.91\%   & +0.99\%   & +1.92\%   & +0.18\%   & +1.01\%   & +2.03\%   \\
 \bottomrule
\end{tabular}  
  }
\end{table*}

\begin{table}[htbp]
\centering
\caption{Performance comparison (\%) of different methods on large datasets. The best results are in bold, and the second-best results are underlined.}
\resizebox{0.45\textwidth}{!}{
\begin{tabular}{cccc}
\toprule
Model & Ogbn-proteins & Amazon2m & Pokec \\
\midrule
\# nodes & 132,534 & 2,449,029 & 1,632,803 \\
\# edges & 39,561,252 & 61,859,140 & 30,622,564 \\
\midrule
MLP & 72.04 $\pm$ 0.48 & 63.46 $\pm$ 0.10 & 60.15 $\pm$ 0.03 \\
GCN & 72.51 $\pm$ 0.35 & 83.90 $\pm$ 0.10 & 62.31 $\pm$ 1.13 \\
SGC & 70.31 $\pm$ 0.23 & 81.21 $\pm$ 0.12 & 52.03 $\pm$ 0.84 \\
GCN-NSampler & 73.51 $\pm$ 1.31 & 83.84 $\pm$ 0.42 & 63.75 $\pm$ 0.77 \\
GAT-NSampler & 74.63 $\pm$ 1.24 & 85.17 $\pm$ 0.32 & 62.32 $\pm$ 0.65 \\
SIGN & 71.24 $\pm$ 0.46 & 80.98 $\pm$ 0.31 & 68.01 $\pm$ 0.25 \\
NodeFormer & 77.45 $\pm$ 1.15 & 87.85 $\pm$ 0.24 & 70.32 $\pm$ 0.45 \\
SGFormer & \underline{79.53 $\pm$ 0.38} & \underline{89.09 $\pm$ 0.10} & \underline{73.76 $\pm$ 0.24} \\
\rowcolor[gray]{0.9}
VecFormer &\textbf{80.52 $\pm$ 0.20} & \textbf{89.72 $\pm$ 0.04} &  \textbf{78.06 $\pm$ 0.21}  \\
\bottomrule
\end{tabular}
}
\label{large_dataset}
\end{table}

\subsection{Node Classification Task (RQ1)}

\textbf{Setup.} We evaluated VecFormer on node classification task using widely-used graph datasets. These include six homophilic graphs 
\cite{DBLP:journals/aim/SenNBGGE08,shchur2019pitfallsgraphneuralnetwork}
: CoraFull, Pubmed, Computer, Photo, CS, and Physics, as well as two heterophilic graphs \cite{pei2020geomgcngeometricgraphconvolutional}, Chameleon and Squirrel, where redundant nodes were removed, as proposed in the benchmark \cite{platonov2024criticallookevaluationgnns}. Following the settings of prior works, we split the datasets into training, validation, and test sets in the ratio of 60\%, 20\%, and 20\%, respectively. The mean and standard deviation of Accuracy were reported over ten runs. Detailed dataset information can be found in Appendix \ref{dataset_info}.

\textbf{Baselines.} We compared VecFormer with three categories of GNN methods: standard GNNs, including GCN \cite{kipf2017semisupervisedclassificationgraphconvolutional}, GAT \cite{veličković2018graphattentionnetworks}, and GraphSAGE \cite{hamilton2018inductiverepresentationlearninglarge} ; recent Graph Transformer models, such as GOAT \cite{kong2023goat}, GraphGPS \cite{rampášek2023recipegeneralpowerfulscalable}, NodeFormer \cite{wu2023nodeformerscalablegraphstructure}, SGFormer \cite{wu2024sgformersimplifyingempoweringtransformers}, Exphormer \cite{shirzad2023exphormersparsetransformersgraphs}, NAGphormer \cite{chen2023nagphormertokenizedgraphtransformer}, and Polynormer \cite{deng2024polynormerpolynomialexpressivegraphtransformer}; and vector quantization-based graph distillation methods, exemplified by VQGraph \cite{yang2024vqgraphrethinkinggraphrepresentation}. For VecFormer, we use GAT as the GNN encoder.

\begin{figure}[htbp]
    \centering
    \includegraphics[width=0.45\textwidth]{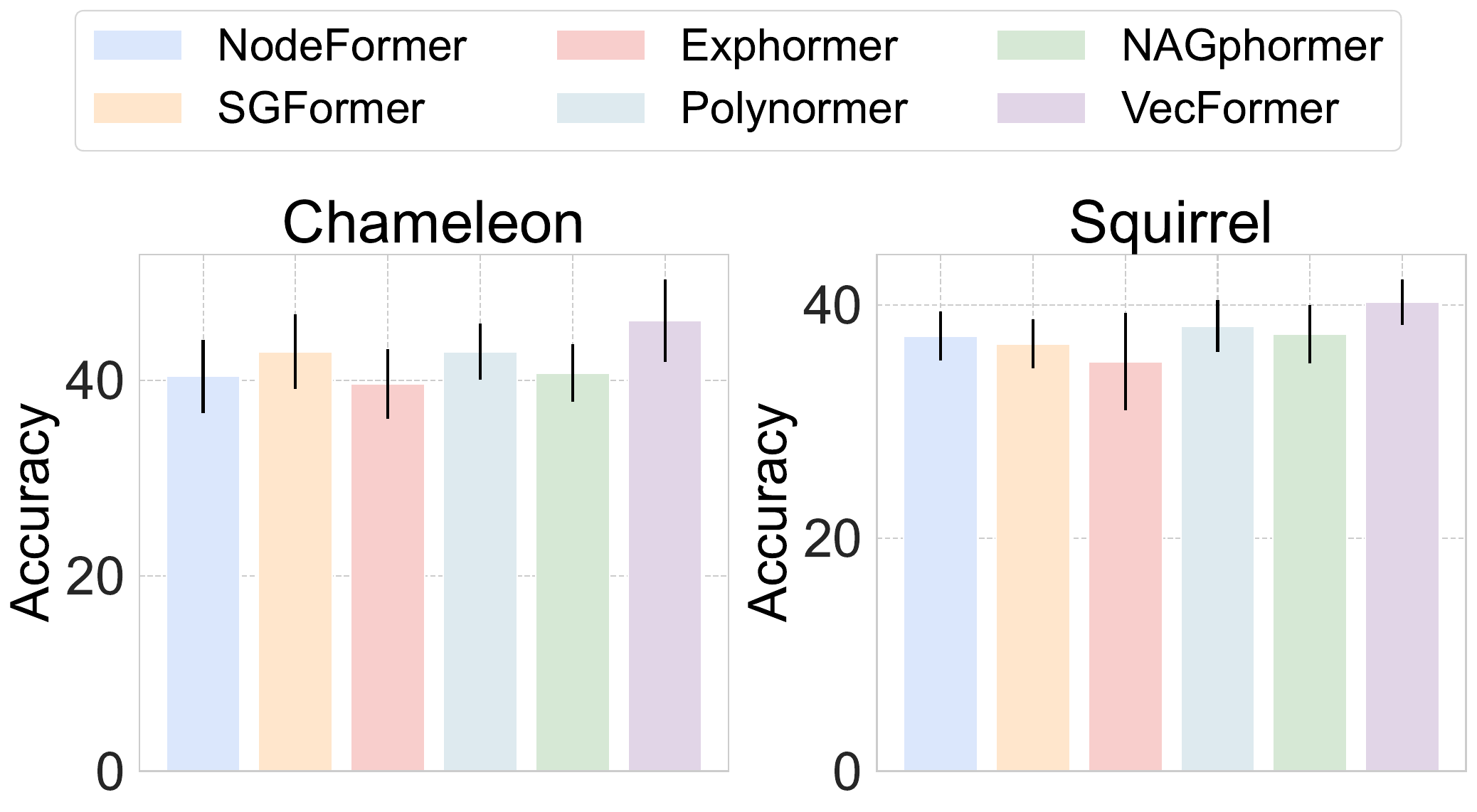} 
    \caption{Experimental results(\%) for node classification task on heterophily graph datasets Chameleon and Squirrel.}
    \label{hetero} 
\end{figure}

\textbf{Results.} The node classification results on homophilic graphs are presented in Table \ref{homo_dataset}, while results on heterophilic graphs are shown in Figure \ref{hetero}. VecFormer outperforms standard GNNs, Graph Transformer models, and the graph distillation method VQGraph on the node classification task. Notably, VecFormer achieved greater improvements on heterophilic graphs compared to homophilic graphs. Compared to the homophilic graphs, which neighboring nodes often share the same label as the central node, the test sets of heterophilic graphs typically exhibit neighborhood structures different from those in the training set. Consequently, Graph Transformers with strong generalization and representational capabilities tend to perform better. VecFormer, along with Polynormer, which possesses polynomial-expressive, demonstrated outstanding performance on the two heterophilic graph datasets.

\subsection{OOD Generalization Scenario (RQ2)}
\textbf{Setup.} We select two types of OOD datasets to evaluate the effectiveness of VecFormer in OOD scenarios. Following the setup of CaNet \cite{wu2024graphoutofdistributiongeneralizationcausal}, we synthetically generate spurious node features to introduce distribution shifts in the Cora and Citeseer datasets. Additionally, from the perspective of graph structure, we select subgraph-level partitioned datasets, as each subgraph has different sizes, densities, and node degrees. For each dataset, we divide the ID dataset into training, validation, and test sets with ratios of 50\%, 25\%, and 25\%, respectively. The trained model is then tested on both ID test set and the OOD set. Each experiment is run five times, and we report the accuracy on the Cora and Citeseer OOD datasets and the ROC AUC on the Twitch dataset.

\textbf{Baselines.} We select two types of methods as baselines for comparison: GT and graph OOD methods such as ERM \cite{788640}, IRM \cite{arjovsky2020invariantriskminimization}, Coral \cite{sun2016deepcoralcorrelationalignment}, DANN \cite{ganin2016domainadversarialtrainingneuralnetworks}, GroupDRO \cite{sagawa2020distributionallyrobustneuralnetworks}, Mixup \cite{zhang2018mixupempiricalriskminimization}, SR-GNN \cite{zhu2021shiftrobustgnnsovercominglimitations}, EERM \cite{wu2024handlingdistributionshiftsgraphs}, and CaNet \cite{wu2024graphoutofdistributiongeneralizationcausal} (\textbf{the SOTA Graph OOD method}). We use the GAT as the encoder backbone.

\begin{figure}[htbp]
    \centering
    \includegraphics[width=0.45\textwidth]{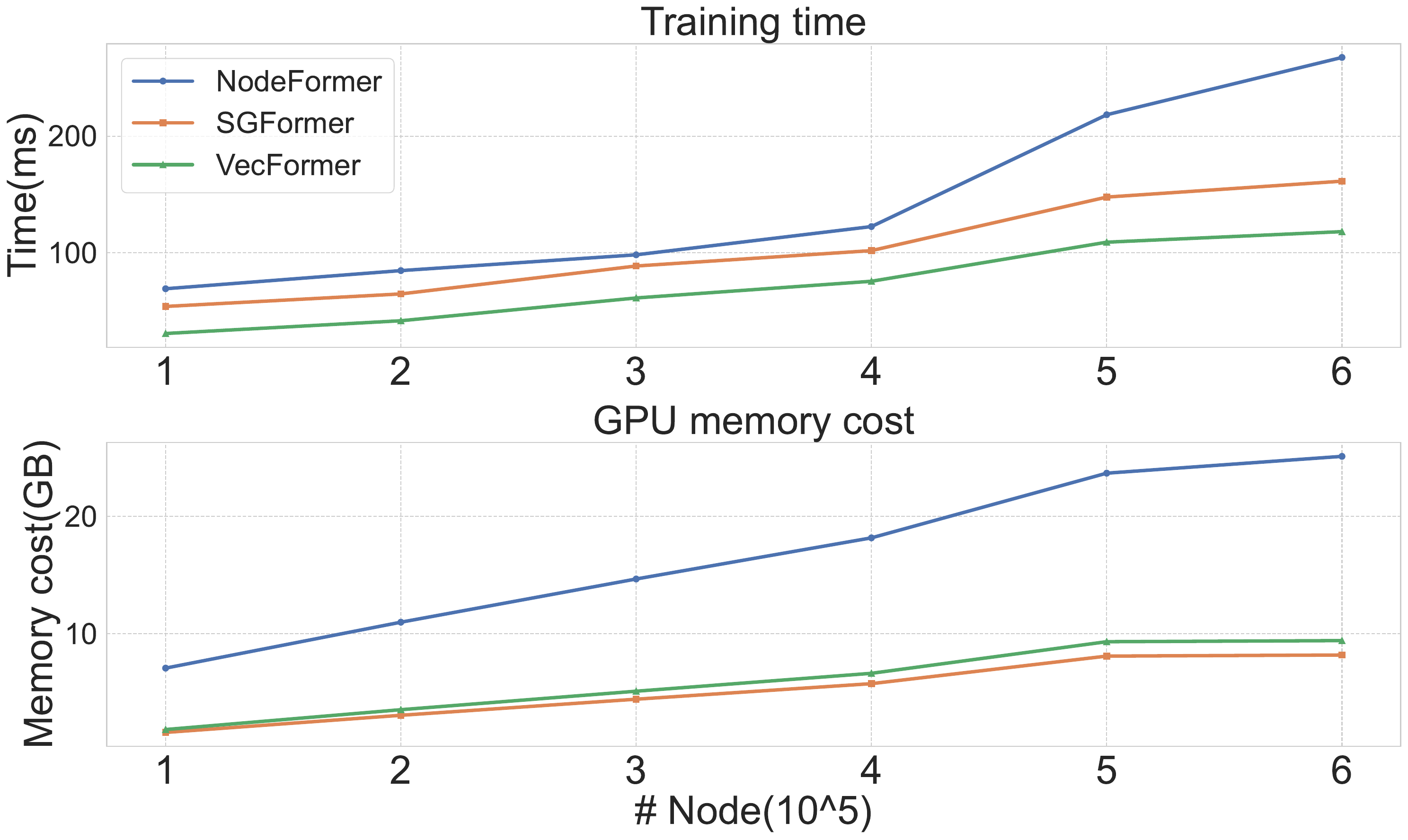} 
    \caption{The training time per epoch and GPU memory usage of three linear-complexity graph transformers on graphs with varying numbers of nodes.}
    \label{compute_and_mem} 
\end{figure}


\textbf{Results.} From Table \ref{ood_result}, we can observe that: (1) Graph Transformers perform better on ID data but underperform on OOD data compared to Graph OOD generalization methods, due to overfitting on ID data. (2) Graph Transformers perform worse on node distribution shift OOD datasets, as the significant difference between OOD and ID node features validates the inference in Section \ref{model_analysis}. (3) On all ID and OOD test sets VecFormer outperforms the SOTA Graph OOD generalization method, CaNet. Moreover, compared to the best-performing Graph Transformer, VecFormer achieves improvements of \textbf{8.67\%} and \textbf{16.91\%} on the Cora  OOD and Citeseer OOD test sets, demonstrating the generalization capability of graph token attention.



\subsection{Performance on Large-sized Graph (RQ3)}

\textbf{Setup.} We evaluate VecFormer from three aspects: model performance, training time, and GPU memory consumption. We assess VecFormer's performance on large graphs with sizes ranging from 0.1M to 2M, including Ogbn-proteins \cite{hu2021opengraphbenchmarkdatasets}, Pokec 
\cite{Leskovec_Krevl_2014}
, and Amazon2m \cite{Chiang_2019}. The dataset partitioning follows the settings in the SGFormer \cite{wu2024sgformersimplifyingempoweringtransformers}. To verify the scalability of the model, we sample subgraphs with node sizes ranging from 100K to 600K from Amazon2m as datasets, and record the training time and GPU memory consumption per epoch.

\textbf{Baselines.} We select standard GNNs, GCN and GAT, sampling-based GNNs \cite{zeng2020graphsaintgraphsamplingbased}, GCN-NSampler and GAT-NSampler, scalable GNNs, SGC \cite{wu2019simplifyinggraphconvolutionalnetworks} and SIGN \cite{frasca2020signscalableinceptiongraph},  and two representative linear-complexity Graph Transformers, NodeFormer \cite{wu2023nodeformerscalablegraphstructure} and SGFormer \cite{wu2024sgformersimplifyingempoweringtransformers}, as baselines.

\textbf{Results.} From Figure \ref{compute_and_mem}, we observe that the training time and GPU memory consumption of the three linear-time-complexity Graph Transformers differ. VecFormer and SGFormer have significantly lower training time and memory consumption compared to NodeFormer. VecFormer has a shorter training time than SGFormer, with slightly higher memory consumption. In terms of model performance, as shown in Table \ref{large_dataset}, VecFormer significantly outperforms the other two Graph Transformers on all three datasets. From the three evaluation perspectives, VecFormer demonstrates superior model performance and training time compared to NodeFormer and SGFormer, proving the scalability of graph token attention.

\subsection{Effectiveness Analysis (RQ4)}
In this section, we analyze the effectiveness of VecFormer through \texttt{Graph Codes} and \texttt{Graph Tokens}.

\textbf{Graph Codes.} As the fundamental unit of vector quantization, \texttt{Graph Codes} influence the Graph Codebook Training phase. We set the sizes of the feature codebook and structure codebook to 2, 4, 8, 16, and 32, and record the values of various reconstruction losses during training. From Figure \ref{rec_loss}, we observe that as the number of codes in the codebooks increases, reconstruction losses gradually decrease until stabilize.

\textbf{Graph Tokens.} The \texttt{Graph Token List} serves to compress global graph information, enabling nodes to interact with it through cross-attention mechanisms to capture global context information. As shown in Figure \ref{token_number}, when the size of the \texttt{Graph Token List} is set to 4, 16, 64, or 256, the model performs poorly with a smaller number of \texttt{Graph Tokens}, as it struggles to effectively capture and compress graph information. As the number of \texttt{Graph Tokens} increases, the model's performance gradually improves and stabilizes. In the meantime, to verify whether VecFormer learns beneficial attention patterns, we visualize dimensionality-reduced attention weights (rather than model outputs) between nodes and \texttt{Graph Token List} on OOD datasets. As can be observed from Figure \ref{tsne}, the attention weights correlate with labels and exhibit non-uniform distributions, proving that VecFormer learns meaningful contextual patterns.

\section{Conclusion}

In this work, we focus on improving the GT's computational efficiency and performance in OOD scenarios. We introduce a variant of VQ, SoftVQ, which is applied to GT to compress the distribution on graphs to reduce computational complexity. In the meantime, both theoretical and empirical evidence demonstrate the superiority of VecFormer in OOD scenarios. We believe this paper will open new directions for designing more efficient and generalizable Graph Transformers. Interesting directions for future work include: (1) exploring GT in graph imbalance scenarios; and (2) designing graph foundation models based on vector quantization.

\section*{Acknowledgments}
This work was supported by National Science and Technology Innovation - Major Program(No. 2022ZD0115101), the National Science and Technology Major Project of China (No.2021YFA1301603), and the Project (No. WU2025B006) from the SOE Dean Special Project Fund (SOE-DSPF) Program of Westlake University.

\bibliographystyle{ACM-Reference-Format}

\clearpage
\appendix

\section{Design Details \& Hyper-parameter Settings}

In this section, we introduce the design details of our proposed method and summarize the hyper-parameter settings. In VecFormer, a GAT with residual connections is employed as the GNN encoder, while the number of cross-attention layers is set to 1. The Feature Decoder, Structure Decoder, and Graph Decoder are all single-layer MLPs. The training hyperparameters of VecFormer are selected using grid search with the following search space: learning rate \{0.001, 0.005, 0.01\}, hidden size \{64, 128, 256\}, weight decay \{1e-3, 5e-4, 1e-4\}, and dropout ratio \{0.1, 0.3, 0.5, 0.7\}. The hyperparameters related to the number of codes and tokens are shown in Table \ref{token_num_info}. 

\begin{table}[H]
\centering
\caption{Hyper-parameter settings of our proposed method.}
\begin{tabular}{lcccc}
    \toprule
    Dataset & \makecell{Feature \\ Codebook \\ Size} & \makecell{Structure \\ Codebook \\ Size} & $N_f$ & $N_s$ \\ 
    \midrule
    Pubmed        & 256 & 256 & 64 & 64\\
    CoraFull      & 256 & 256 & 64 & 64\\
    Computer      & 256 & 256 & 64 & 64\\
    Photo         & 256 & 256 & 64 & 64\\
    CS            & 256 & 256 & 64 & 64\\
    Physics       & 256 & 256 & 64 & 64\\
    Chameleon     & 256 & 256 & 32 & 32\\
    Squirrel      & 128 & 128 & 32 & 32\\
    \midrule
    Cora OOD      & 64 & 64 & 16 & 16\\
    Citeseer OOD  & 64 & 64 & 16 & 16\\
    Twitch        & 512 & 512 & 64 & 64\\
    \midrule
    Ogbn-proteins & 64  & 64  & 16 & 16 \\
    Amazon2m      & 256 & 256  & 64 & 64\\
    Pokec         & 256 & 256  & 64 & 64\\
    \bottomrule
\end{tabular}

\label{token_num_info}
\end{table}

\section{Datasets}
\label{dataset_info}
\begin{table}[h!]
\caption{The statistical information of eight datasets.}
\centering
\begin{tabular}{lcccc}
\toprule
\textbf{Dataset} & \textbf{\# Nodes} & \textbf{\# Edges} & \textbf{\# Feature Dims} & \textbf{\# Classes} \\ 
\midrule
Pubmed          & 19,717            & 44,324            & 500                      & 3                  \\
CoraFull        & 19,793            & 126,842           & 8,710                    & 70                 \\
Computer        & 13,752  & 245,861& 767& 10\\
Photo           & 7,650&  119,081& 745& 8\\
CS              & 18,333            & 81,894           & 6,805                    & 15                 \\
Physics         & 34,493            & 247,962           & 8,415                    & 5                  \\
Chameleon       &      890 &8,854 &  2,325 & 5        \\
Squirrel       &    2,223 &  46,998 & 2,089 & 5    \\
\bottomrule
\end{tabular}
\label{tab:dataset_info}
\end{table}

\section{Application in Gene Perturbation Prediction}
In recent years, graph neural networks have demonstrated strong capability in modeling complex relational structures, making them particularly suitable for biological network scenarios where high-dimensional dependencies can be represented as nodes and edges. Regulatory and co-expression relationships among genes naturally form graph structures, and gene perturbations can trigger cascading effects over the network, ultimately manifested as a set of differentially expressed (DE) genes. Therefore, we formulate gene-perturbation response prediction as a node classification task on a gene co-expression network: given a perturbation condition, we predict for each gene whether it is differentially expressed, thereby leveraging GNNs' strengths in graph information propagation and representation learning to better model perturbation effects.

\textbf{Dataset and Task.}
We evaluate our method on the Virtual Cell Challenge (VCC) \cite{roohani2025virtual} training dataset, which contains single-cell RNA-seq data from 221,273 cells and 18,080 genes. The dataset includes 150 gene-perturbation (gene knockout) experiments and 38,176 non-targeting control cells. We model the problem as a node classification task on a gene co-expression network, aiming to predict DE genes for each perturbation experiment.

\textbf{Data Split.}
Using a fixed random seed, we randomly split the 150 perturbations into a training set (80), a validation set (20), and a test set (50). The validation set is used for model selection (checkpoint selection), and the test set is used only for final evaluation.

\textbf{Graph Construction.}
We construct a symmetric k-nearest-neighbor (k-NN) co-expression network based on Pearson correlation coefficients computed from the control expression matrix. We experiment with $k \in \{5, 10, 20\}$ to study the effect of graph density on model performance.

\textbf{Node Features.}
For each perturbation experiment, node features are constructed as follows: (1) we apply PCA \cite{mackiewicz1993principal} to the control gene expression matrix (gene $\times$ cell) to reduce the dimension to 128, obtaining a base representation for each gene; (2) for a specific perturbation targeting gene $g$, we add $g$'s PCA embedding to the representations of all genes, enabling the model to perceive which gene is perturbed.

\textbf{Label Generation.}
We perform the Wilcoxon rank-sum test to compare each gene's expression between perturbed cells and control cells, followed by Benjamini--Hochberg FDR correction. Genes with adjusted $p$-values $< 0.05$ are labeled as differentially expressed. We consider a binary classification setting.

\textbf{Evaluation Metric.}
We report the Differential Expression Score (DES): following the official VCC metric, DES measures the overlap between the predicted DE gene set and the ground-truth DE gene set, normalized by the number of true DE genes. When the predicted set is larger than the true set, we select the top-$k$ genes with the highest predicted DE probabilities.

\textbf{Baselines and Implementation.}
We compare four models: GCN, GAT, and VecFormer (our method). All models use 2 layers with hidden dimension 128 and dropout rate 0.2. VecFormer is trained in two stages: a pretraining stage with a reconstruction loss for 100 epochs, and a classification stage trained for up to 300 epochs with early stopping based on validation performance.

\textbf{Results.} Table~\ref{tab:des_knn} reports the Differential Expression Score (DES) under different $k$-NN graph densities. Overall, VecFormer consistently improves over its corresponding GNN backbone across all graph settings, demonstrating the effectiveness of the proposed perturbation-aware representation learning.

\begin{figure*}[h]
    \centering
    \includegraphics[width=0.9\textwidth]{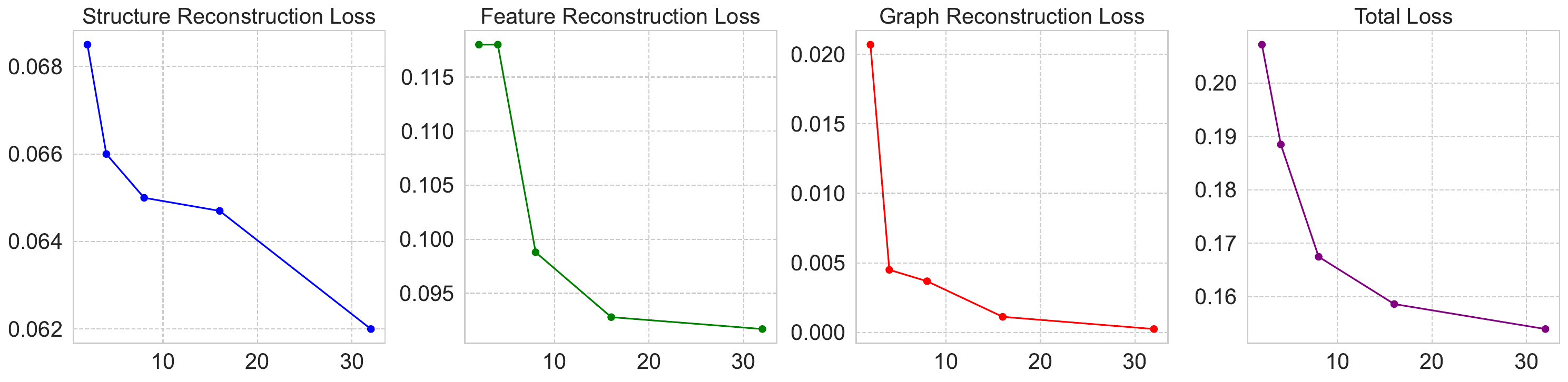} 
    \caption{Reconstruction losses of different types and the total loss under varying numbers of \texttt{Graph Codes}.}
    \label{rec_loss} 
\end{figure*}

\begin{table*}[tbp]
\centering
\caption{DES (mean$\pm$std) on the VCC test set under different $k$-NN graph densities. We additionally report the mean improvement of VecFormer over its corresponding backbone: $\Delta$DES (absolute gain in mean DES) with relative improvement in parentheses.}
\label{tab:des_knn}
\begin{tabular}{lccc}
\toprule
Model & $k=5$ & $k=10$ & $k=20$ \\
\midrule
GCN & $0.1286\pm0.1462$ & $0.0085\pm0.0420$ & $0.0031\pm0.0148$ \\
\rowcolor[gray]{0.9}VecFormer \underline{(GCN backbone)} & $\textbf{0.1490}\pm\textbf{0.2251}$ & $\textbf{0.0699}\pm\textbf{0.1566}$ & $\textbf{0.1000}\pm\textbf{0.1989}$ \\
$\Delta$DES vs.\ GCN & $+0.0204\;(+15.9\%)$ & $+0.0614\;(+722.4\%)$ & $+0.0969\;(+3125.8\%)$ \\
\midrule
GAT & $0.0316\pm0.0197$ & $0.0244\pm0.0148$ & $0.0445\pm0.0358$ \\
\rowcolor[gray]{0.9}VecFormer \underline{(GAT backbone)} & $\textbf{0.1322}\pm\textbf{0.2226}$ & $\textbf{0.1298}\pm\textbf{0.2045}$ & $\textbf{0.1445}\pm\textbf{0.1829}$ \\
$\Delta$DES vs.\ GAT & $+0.1006\;(+318.4\%)$ & $+0.1054\;(+432.8\%)$ & $+0.1000\;(+224.7\%)$ \\
\bottomrule
\end{tabular}
\end{table*}

For the GCN backbone, VecFormer achieves higher mean DES for all $k$. The gain is modest on the sparse graph ($k=5$), improving from 0.1286 to 0.1490 ($\Delta$DES=+0.0204, +15.9\%). As the graph becomes denser, the improvement becomes substantially larger: at $k=10$, VecFormer increases DES from 0.0085 to 0.0699 (+0.0614, +722.4\%), and at $k=20$ from 0.0031 to 0.1000 (+0.0969, +3125.8\%). These results suggest that GCN is sensitive to graph density, while VecFormer benefits from richer neighborhood structure.

A similar trend is observed with the GAT backbone. VecFormer yields consistent improvements over GAT across all densities: from 0.0316 to 0.1322 at $k=5$ (+0.1006, +318.4\%), from 0.0244 to 0.1298 at $k=10$ (+0.1054, +432.8\%), and from 0.0445 to 0.1445 at $k=20$ (+0.1000, +224.7\%). Notably, VecFormer with a GAT backbone attains the best performance, peaking at 0.1445 DES when $k=20$.

In summary, VecFormer provides substantial accuracy gains on DE gene prediction and exhibits strong robustness to graph construction choices, indicating improved capability in capturing perturbation-induced effects on gene co-expression networks.

\section{Effectiveness Analysis}

\begin{figure}[H]
    \centering
    \includegraphics[width=0.45\textwidth]{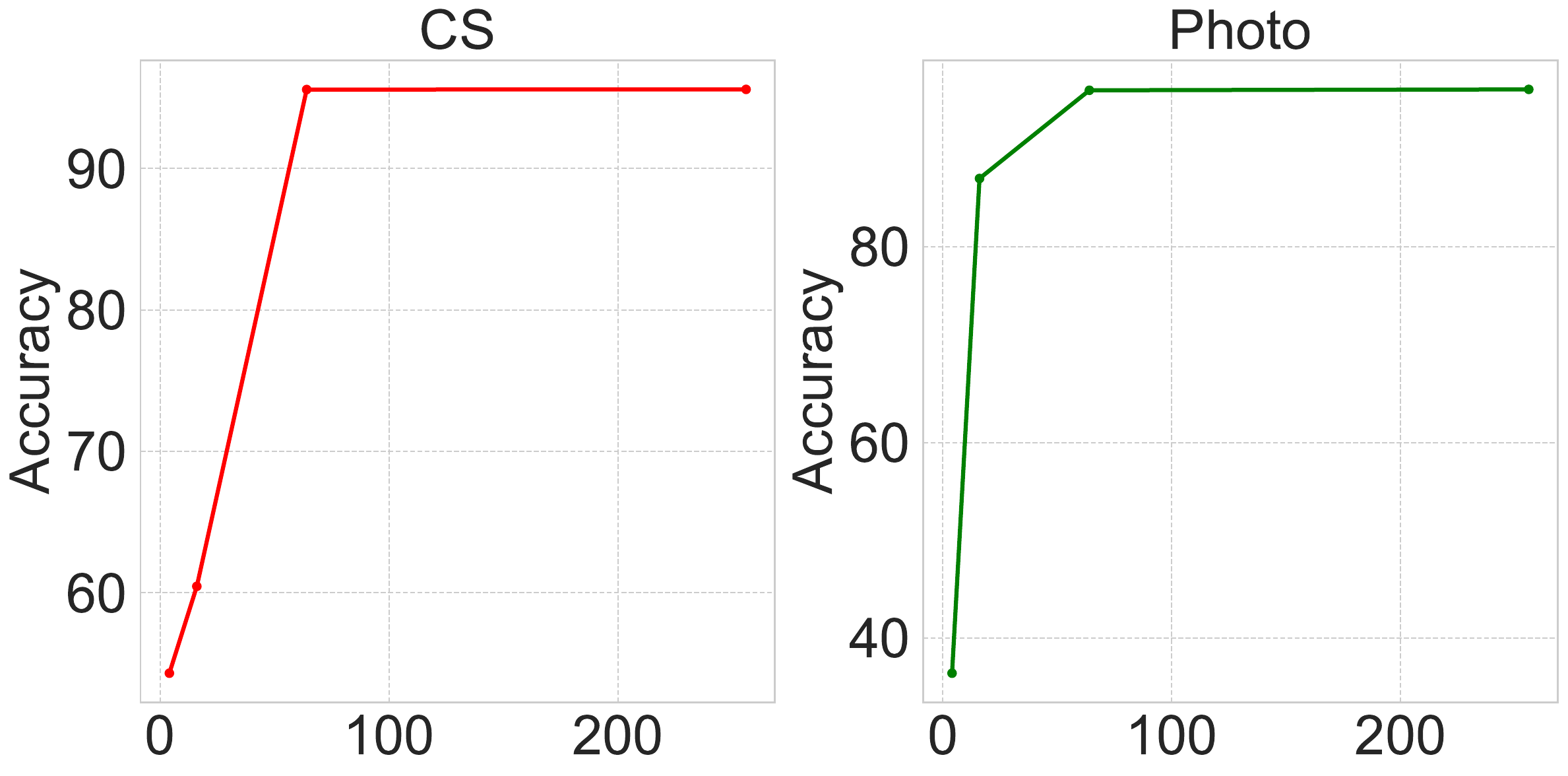} 
    \caption{VecFormer's performance(\%) under varying numbers  of \texttt{Graph Tokens}.}
    \label{token_number} 
\end{figure}

\begin{figure}[H]
    \centering
    \includegraphics[width=0.45\textwidth]{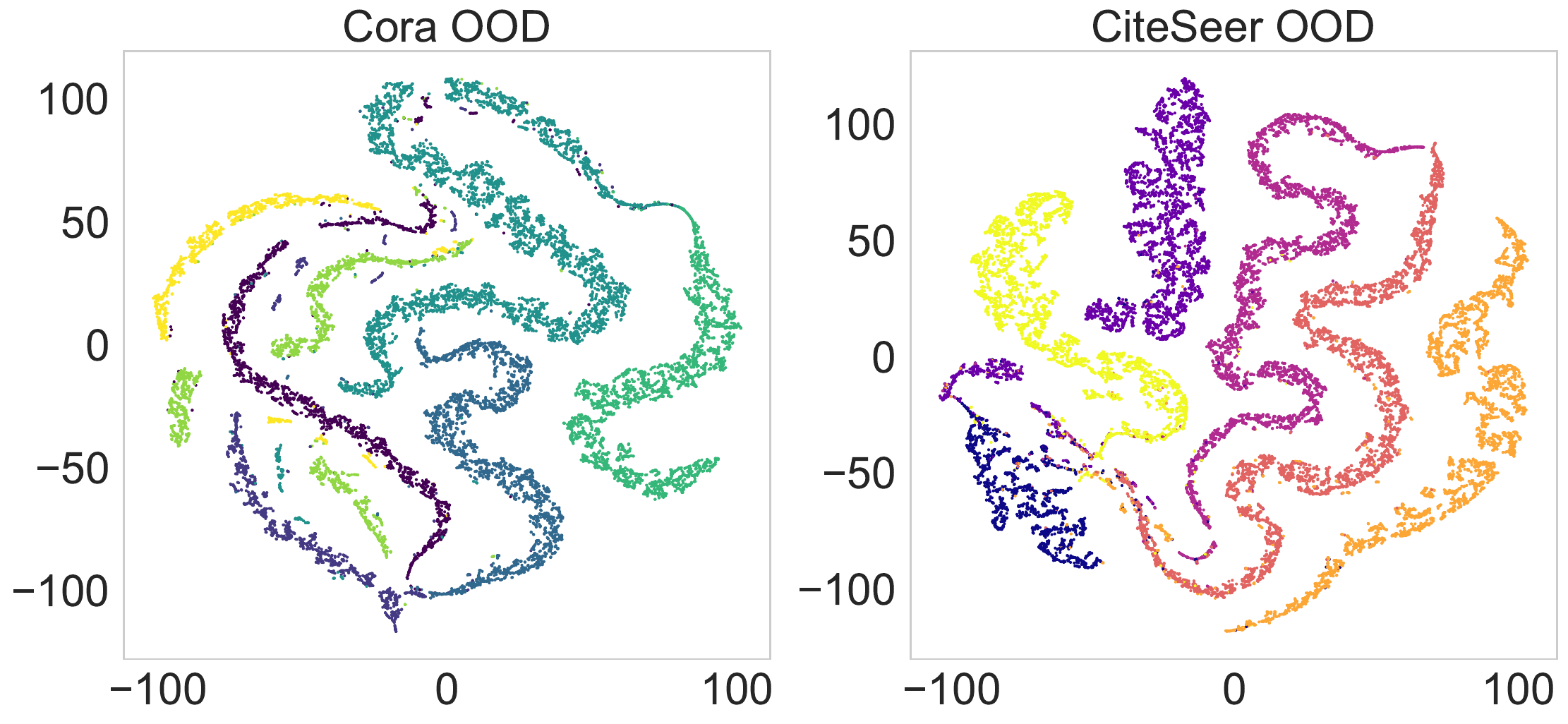} 
    \caption{$t$-SNE visualization of attention weights on OOD datasets. }
    \label{tsne} 
\end{figure}

\end{document}